# 1D-CapsNet-LSTM: A Deep Learning-Based Model for Multi-Step Stock Index Forecasting

**Cheng Zhang** [1,*], **Nilam Nur Amir Sjarif** [1], **Roslina Ibrahim** [1]

[1] Advanced Informatics Department, Razak Faculty of Technology and Informatics, Universiti Teknologi Malaysia, 54100, Kuala Lumpur, Malaysia

* Corresponding author. Tel: +60 19-330 0187

*Address*: Advanced Informatics Department, Razak Faculty of Technology and Informatics, Universiti Teknologi Malaysia, 54100, Kuala Lumpur, Malaysia

ORCID:

Cheng Zhang: 0000-0002-4150-3371

Nilam Nur Amir Sjarif: 0000-0003-4969-9708

Roslina Ibrahim: 0000-0002-1343-5842

*Email addresses*:

zcheng582dx@gmail.com (Cheng Zhang)

nilamnur@utm.my (Nilam Nur Amir Sjarif)

iroslina.kl@utm.my (Roslina Ibrahim)

**Abstract**

Multi-step stock index forecasting is vital in finance for informed decision-making. Current forecasting methods on this task frequently produce unsatisfactory results due to the inherent data randomness and instability, thereby underscoring the demand for advanced forecasting models. Given the superiority of capsule network (CapsNet) over CNN in various forecasting and classification tasks, this study investigates the potential of integrating a 1D CapsNet with an LSTM network for multi-step stock index forecasting. To this end, a hybrid 1D-CapsNet-LSTM model is introduced, which utilizes a 1D CapsNet to generate high-level capsules from sequential data and a LSTM network to capture temporal dependencies. To maintain stochastic dependencies over different forecasting horizons, a multi-input multi-output (MIMO) strategy is employed. The model's performance is evaluated on real-world stock market indices, including S&P 500, DJIA, IXIC, and NYSE, and compared to baseline models, including LSTM, RNN, and CNN-LSTM, using metrics such as RMSE, MAE, MAPE, and TIC. The proposed 1D-CapsNet-LSTM model consistently outperforms baseline models in two key aspects. It exhibits significant reductions in forecasting errors compared to baseline models. Furthermore, it displays a slower rate of error increase with lengthening forecast horizons, indicating increased robustness for multi-step forecasting tasks.

## 1. Introduction

Stock market indices serve as vital indicators of financial market health and performance. Accurately forecasting future stock index values is crucial in the financial sector as it aids investors, traders, and financial institutions in making well-informed decisions, managing risks, and optimizing their investment strategies (Cavalcante et al., 2016; Tang et al., 2022). Rather than a one-step forecasting approach, multi-step forecasting, which predicts the price of a target variable at multiple consecutive time steps in the future, provides decision-makers with valuable insights into future price fluctuations over a specific time horizon (Aryal et al., 2020; Duan & Kashima, 2021; Zhang et al., 2023b).

One effective method for multi-step forecasting of stock indices is the Multiple-Input Multiple-Output (MIMO) strategy. This strategy involves generating several consecutive predicted values in a single step, without incurring high computational costs (Bontempi, 2008; Bontempi & Taieb, 2011). The MIMO strategy retains the stochastic dependencies between predicted values and is generally more effective than single-output approaches (Taieb et al.,

2012). It has gained widespread adoption in LSTM-based forecasting models for multi-step time series forecasting tasks (Deng et al., 2022; Nguyen et al., 2021; Zhang et al., 2023b). This is because LSTM layers are particularly suitable for capturing temporal dependencies within data and are a preferred choice for building time series forecasting models (Durairaj & Mohan, 2019; Hu et al., 2021; Kumar et al., 2021; Lara-Benítez et al., 2021; Nosratabadi et al., 2020; Sezer et al., 2020).

However, multi-step forecasting of stock indices using LSTM networks often yields unsatisfactory results due to the stochastic and volatile nature of the data. The Efficient Market Hypothesis (EMH) posits that asset prices reflect all available information (Fama, 1970), and in an efficient market, price behavior resembles a random walk, making it difficult to discern patterns in historical data (Fama, 1995). Stock market indices share characteristics with most asset price series, leading to a rapid deterioration in forecasting accuracy as the forecasting horizon extends. Ongoing challenges in this area involve improving forecasting accuracy and understanding the reasons behind such improvements (Makridakis et al., 2018).

One approach to enhancing multi-step forecasting accuracy is to integrate time-frequency decomposition technologies, such as multivariate empirical mode decomposition (MEMD) and multivariate variational mode decomposition (MVMD), into LSTM-based forecasting models. These techniques decompose stock index series into a set of intrinsic mode functions (IMFs), which are then input to the LSTM model for prediction (Deng et al., 2022; Zhang et al., 2023b). However, while decomposition techniques effectively capture time series components like trends, seasonality, and residuals, they incorporate out-of-sample data into the training data, leading to overly optimistic forecasting accuracy (Hewamalage et al., 2023; Wu et al., 2022; Zhang et al., 2023a; Zhang et al., 2015). Nevertheless, these techniques underscore the importance of sophisticated feature extraction, mitigating the drawbacks of an LSTM-based forecasting model with a univariate input (Altan et al., 2021). Proper feature extraction expands the model's input to be multivariate, which can reveal patterns or relationships not readily apparent in raw data, enhancing the learning capability of the forecasting model.

Another way to boost feature extraction is by using a neural network capable of automatically extracting features from univariate raw data. The convolutional neural network long short-term memory (CNN-LSTM) network, for example, leverages a 1D CNN to automatically extract spatial patterns in time series data and is widely used for forecasting financial time series prices (Aldhyani & Alzahrani, 2022; Livieris et al., 2020b; Lu et al., 2020). In alignment with this approach, the model performance on multi-step forecasting tasks can potentially be enhanced by implementing a more advanced feature extraction component, such as a capsule network (CapsNet) (Sabour et al., 2017), which has demonstrated advantages over CNNs in image classification (Choudhary et al., 2023; Pawan & Rajan, 2022). CapsNet, when employed as an advanced feature extraction component, holds the potential to improve the performance of LSTM-based time series forecasting models, as it has been combined with LSTM networks for various forecasting tasks (Ma et al., 2021; Qin et al., 2022). However, CapsNets face limitations when applied to 1D sequential data, preventing them from sequentially extracting features for each data point (Butun et al., 2020; Jayasekara et al., 2019; Liang et al., 2022). As a result, the application of CapsNet in financial time series forecasting is not yet common, and its potential as a feature extraction component in multi-step stock index forecasting models remains unexplored.

To investigate such a potential, our study introduces a novel hybrid model, 1D-CapsNet-LSTM, in which a 1D CapsNet generates high-level features, represented as high-level capsules, for each data point in the univariate sequence and a LSTM layer then captures the temporal dependencies between these high-level features. Under the MIMO approach, the 1D-CapsNet-LSTM model produces multiple forecasts for different forecasting horizons in a single step. To assess the performance of this model for multi-step forecasting of real-world stock indices, we selected four prominent stock market indices—Standard and Poor's 500 index (S&P 500), Dow Jones Industrial Average (DJIA), Nasdaq Composite Index (IXIC), and New York Stock Exchange (NYSE)—for our experiments. We evaluated the model's performance using four metrics: root mean squared error (RMSE), mean absolute error (MAE), mean absolute percentage error (MAPE), and Theil inequality coefficient (TIC). Additionally, we included three baseline models (LSTM, RNN, and CNN-LSTM) for performance comparison.

This study makes two significant contributions to the literature. Firstly, we introduce a novel architecture of 1D CapsNet that can generate high-level capsules for each data point from 1D sequential data. The time-distributed dynamic routing method used in 1D CapsNet at each time step represents a distinct and innovative methodology. Secondly, to the best of our knowledge, this study is the first to apply a hybrid model that combines CapsNet and LSTM to financial time series forecasting, especially for multi-step forecasting tasks. In summary, this study extends the application of CapsNet in the field of time series forecasting.

The remaining sections of this paper are organized as follows. Section 2 discusses related studies on multi-step forecasting strategies, convolutional-recurrent neural networks, and applications of CapsNets in various domains. In

Section 3, we explain the MIMO strategy for multi-step forecasting and provide a detailed explanation of the proposed 1D-CapsNet-LSTM architecture. Section 4 outlines the experimental setup and presents the results of the performance comparison between the proposed model and the baseline models. Finally, Section 5 concludes the paper.

## 2. Related work

This section introduces studies related to multi-step forecasting strategies, convolutional-recurrent neural networks (CRNNs), and applications of CapsNets in different domains. The insights drawn from these studies imply that the emerging use of CapsNets in time series analysis, particularly in financial time series forecasting, presents an intriguing frontier.

### 2.1. Multi-step forecasting strategies

There are approximately five main strategies for generating multi-step forecasts, as outlined by Taieb et al. (2012): recursive, direct, DirRec, MIMO, and DIRMO. The first strategy is the recursive or iterated approach, which refers to training a one-step forecasting model and using the previous time step's forecast as its input for the subsequent forecast (Taieb & Hyndman, 2012). The limitation of this strategy is that the forecasting error accumulates rapidly after a few steps. The second is the direct strategy, which generates multiple forecasts using multiple corresponding models (Cheng et al., 2006). The problem with using the direct strategy is that the computational cost is high and the dependencies among consecutive predicted values are broken. The third is the DirRec strategy, which is a combination of the two previous strategies (Taieb & Hyndman, 2012). Under this strategy, the forecast horizon is divided into several groups, after which the first group of forecasts is generated directly using a set of models, and the following groups of forecasts are recursively produced.

The multiple-input multiple-output (MIMO) strategy was introduced to preserve the stochastic dependencies between predicted values (Bontempi, 2008; Bontempi & Taieb, 2011). Unlike previous approaches, a forecasting model under the MIMO strategy generates the predicted values for all forecasting horizons in a single step. Another strategy, namely, DIRMO, was developed to merge the best aspects of DirRec with those of MIMO (Taieb et al., 2009). DIRMO aims to balance the stochastic dependence between the forecasted values and maintain model flexibility. Although these five forecasting strategies have been described separately in the literature and sometimes with different terminologies, multiple-output approaches are generally more effective than single-output approaches (Taieb et al., 2012).

In the domain of financial time series forecasting, a direct strategy is commonly used to predict a specific value several time steps away from the current time step rather than several values at consecutive future time steps (Lin et al., 2022; Paquet & Soleymani, 2022; Tripathi & Sharma, 2022; Wang & Wang, 2020). In contrast, the MIMO strategy is often employed when multi-step forecasting results are needed (Aryal et al., 2020; Deng et al., 2022; Staffini, 2022).

**2.2. Convolutional-recurrent neural networks**

Convolutional-recurrent neural networks (CRNNs) are a type of hybrid neural network developed by combining the complementary strengths of CNNs and RNNs to capture spatiotemporal patterns in various tasks, including time series forecasting (Shi et al., 2017; Zhan et al., 2018). In a CRNN model for time series forecasting, a 1D CNN extracts spatial features from the input sequence and outputs a sequence of feature vectors, which are then fed into a recurrent layer that captures temporal dependencies and long-term patterns in the feature map. Therefore, the use of a 1D CNN mitigates the drawbacks of a single RNN model with a univariate input. In the domain of forecasting financial time series prices, one variant of CRNN networks, CNN-LSTM, has found extensive applications (Aldhyani & Alzahrani, 2022; Livieris et al., 2020a; Livieris et al., 2020b; Lu et al., 2020). By incorporating a memory mechanism, the LSTM cells can store information for long durations, selectively forget irrelevant information, and update content based on new input, thereby offering the LSTM layer the ability to address the "vanishing gradient" problem of simple RNNs (Hochreiter & Schmidhuber, 1997).

**2.3. CapsNets and their applications**

CapsNets are a type of neural network designed to overcome certain limitations of traditional CNNs in processing hierarchical relationships within data. They improve the recognition of complex patterns by considering the spatial relationships between features in an image (Sabour et al., 2017). Compared with CNNs, which implement elementary pooling that often assigns the same values to adjacent data points in a region within the feature map, possibly discarding useful information, CapsNets use an implementation of the concept of capsules within the feature map and adopt a more complex approach than the pooling operation to process the capsules of data. CapsNets first perform the same convolutions as CNNs and then create primary capsules from the previous convolutional results. Subsequently, high-level capsules in vector form, rather than scalar form, are generated by routing all primary capsules. Through the routing operation, the positional information of the primary capsules is "rate coded" in the real-valued components of the high-level capsules. These high-level capsules, which are groups of neurons that represent specific features, can encode information regarding the presence, orientation, and various properties of a feature, allowing the network to better understand the spatial arrangement of objects within an image. CapsNets have been widely applied in various classification tasks, such as image recognition, pose estimation, and understanding complex visual patterns (Afshar et al., 2018; LaLonde & Bagci, 2018; Ragab et al., 2022; Tampubolon et al., 2019; Xiang et al., 2021).

Moreover, CapsNet and LSTM can be combined into a hybrid neural network, where CapsNet often serves as a metaclassifier that classifies the features learned by LSTM. Such a combination has been adopted for a series of image classification tasks, such as compound fault diagnosis (Ke et al., 2022), emotion recognition (Shahin et al., 2022), and fake news detection (Sridhar et al., 2021). In regression tasks, the CapsNet and LSTM networks can be combined in a manner similar to the structure of CNN-LSTM so that CapsNet is used to sequentially extract high-level capsules from a series of images, and the LSTM layer is then used to generate an estimation of the target variable from these high-level capsules (Ma et al., 2021; Qin et al., 2022). However, when the input is a 1D sequence, CapsNet in the hybrid CapsNet-LSTM neural network cannot sequentially generate high-level capsules corresponding to every data point in the input sequence; therefore, its output is not compatible with the recurrent layer, which limits the application of CapsNet-LSTM networks to time series forecasting tasks.

It is worth noting that the concept "1D CapsNet" has appeared in a few studies. Jayasekara et al. (2019) proposed TimeCaps, which can generate capsules along the temporal dimension to classify electrocardiogram signal beat categories. Butun et al. (2020) used a 1D version of CapsNet for automated coronary artery disease (CAD) detection. Berman (2019) achieved a 1D application of CapsNets for domain generation algorithm (DGA) detection. Similarly, while these studies jointly suggest that CapsNet has potential in time series analysis, these models are incapable of generating high-level capsules corresponding to every data point in the 1D sequence and are therefore incompatible with the recurrent layer for time series forecasting tasks. Overall, the application of CapsNet to financial time series forecasting is not yet common.

## 3. Method

This section introduces the key method and neural architectures adopted in the proposed 1D-CapsNet-LSTM network: the MIMO strategy for generating multi-step forecasts, the 1D CapsNet for feature extraction, and the LSTM network for capturing sequential dependencies within the data. The proposed 1D-CapsNet-LSTM integrates these components to serve as a forecasting model for multi-step stock index forecasting.

## 3.1. MIMO strategy

Given a time series $y$, whose time steps are denoted as $t \in [1, T]$, a multi-step time series forecasting model $f(\cdot)$ under the MIMO strategy estimates in one step several values $[\hat{y}_{t+1}, \cdots, \hat{y}_{t+H}]$ at consecutive $H$ time steps into the future, utilizing the historical values (from time step $t$ to $t-d$, with $d$ being the time lags) of the desired time series, which are denoted as $y_{t-d}, \cdots, y_t$ (Bontempi, 2008; Bontempi & Taieb, 2011). This functional relationship is defined as follows:

$$[\hat{y}_{t+1}, \cdots, \hat{y}_{t+H}] = f([y_{t-d}, \cdots, y_t]; \boldsymbol{\theta}); t > d; H > 1, \tag{1}$$

where vector $\boldsymbol{\theta}$ denotes the model parameters that are adjusted through model training.

The objective of the MIMO approach is to maintain the stochastic dependencies between the predicted values, which helps to avoid the conditional independence assumption made by the direct approach and prevents the accumulation of errors associated with the recursive approach. By generating a vector of future values in one step instead of returning a scalar value, MIMO ensures that the correlations between future observations are captured during model training and utilized in the forecasting process.

## 3.2. 1D CapsNet

A CapsNet with a shallow architecture for image classification typically includes convolutional, capsule, and dynamic routing layers (Sabour et al., 2017). During the dynamic routing process, all primary capsules from the capsule layer are routed to generate high-level capsules. For the classification task, the number of high-level capsules is equal to the number of image categories. In the context of time series forecasting using an LSTM-based model, a feature extraction component should be able to extract features with respect to each data point in the input sequence so that the extracted feature map is compatible with the recurrent layer that produces the final prediction. To provide CapsNets with such feature extraction capabilities, we adopted a different approach to process the 1D sequence extracted from a time series. We assume that, for time series forecasting, each data point in the input sequence corresponds to one high-level capsule. Accordingly, the number of high-level capsules is equal to the sequence length. Another assumption is that a high-level capsule should be produced from a corresponding temporal slice of the primary capsules instead of being generated from all primary capsules. It should be mainly affected by the primary capsules corresponding to the same time step and not by those corresponding to other time steps. Therefore, there should be a "one-to-one correspondence" relationship between the original data points from the input sequence and high-level capsules. Consequently, the routing operation must be applied sequentially to each temporal slice of the primary capsules.

Figure 1 shows a visual representation of the 1D CapsNet for 1D sequence processing. The 1D CapsNet includes a convolutional 1D layer, capsule layer, and time-distributed routing layer. Suppose the convolutional 1D layer has $N$ filters with a size of two, a stride of one, and rectified linear unit (ReLU) activation. Ignoring the batch size, the 1D input sequence through this convolutional layer is converted into a feature map $\mathbf{X} \in \mathbb{R}^{d \times N}$, where $d$ is the sequence length. Given the feature map $\mathbf{X}$, the capsule layer constructs the primary capsules through the following steps. Suppose the primary capsules are 8-element vectors that are oriented along the channel axis, each of which contains the lowest level of features extracted from previous convolutions; then, the feature map $\mathbf{X}$ is divided into $n = N/8$ groups along the channel dimension, reshaped into a new tensor $\mathbf{X}' \in \mathbb{R}^{d \times n \times 8}$ that contains $d \times n$ primary capsules.

All primary capsules are then fed to a "squashing" function to make good use of the nonlinearity. The squashing function is given by Equation (2):

$$\mathbf{v}_{it} = \frac{\|\mathbf{s}_{it}\|^2}{1 + \|\mathbf{s}_{it}\|^2} \frac{\mathbf{s}_{it}}{\|\mathbf{s}_{it}\|}, \qquad (2)$$

where $\mathbf{s}_{it}$ and $\mathbf{v}_{it}$ represent the $i$ th primary capsule from the temporal slice $t$ before and after squashing, respectively. Subsequently, the primary capsule $\mathbf{v}_{it}$ is transformed into a new vector $\mathbf{u}_{it}$ using the transformation matrix $\mathbf{W}_{it}$ such that $\mathbf{u}_{it}$ has the same shape as the high-level capsule $\mathbf{x}_t$, which has a predefined number of elements. The transformation function is given by Equation (3):

$$\mathbf{u}_{it} = \mathbf{W}_{it} \cdot \mathbf{v}_{it} \qquad (3)$$

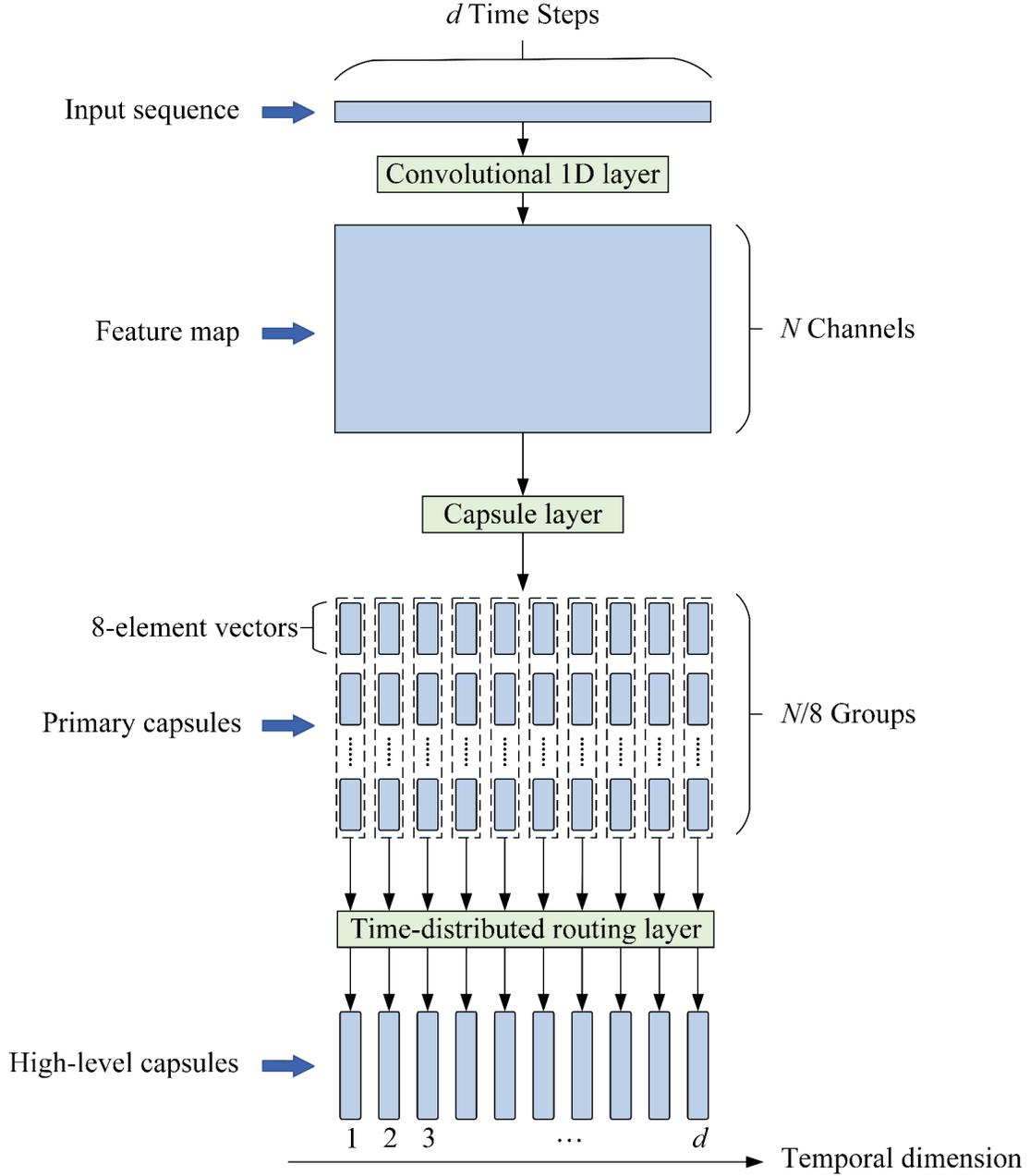

**Figure 1. The 1D CapsNet for 1D sequence processing.**

After matrix transformation, all transformed primary capsules from the temporal slice $t$, denoted as $\mathbf{u}_{it}, i=1,2,\cdots,n,$, are routed to produce one high-level capsule $\mathbf{x}_t$. The overall procedure of the time-distributed dynamic routing is presented in **Algorithm 1**. For each iteration, the initial coupling coefficients $b_{it}, i=1,2,\cdots,n,$ with initial values of zero are converted into coupling coefficients $c_{it}, i=1,2,\cdots,n,$ using a softmax function given by Equation (4), such that the sum of the coupling coefficients $c_{it}$ for temporal slice $t$ is equal to one. Subsequently, a high-level capsule $\mathbf{x}_t$ is generated through the weighted sum of $\mathbf{u}_{it}$ with the coupling coefficients $c_{it}$. The degree of "agreement," or the dot product between $\mathbf{u}_{it}$ and $\mathbf{x}_t$, is added to $b_{it}$ to update $\mathbf{x}_t$ in the following iteration.

$$c_{it} = \frac{\exp(b_{it})}{\sum_i \exp(b_{it})} \tag{4}$$

---

**Algorithm 1.** Time-distributed dynamic routing.

---

1: **procedure** ROUTING $(\mathbf{u}_{it}, d, n, r)$

2:     $\mathbf{u}_{it} \leftarrow$ The $i$ th transformed primary capsule from temporal slice $t$

3:     $d \leftarrow$ The input sequence length

4:     $n \leftarrow$ The number of primary capsules from each temporal slice

5:     $r \leftarrow$ The iteration times of dynamic routing

6:     **for** $t = 1:d$ **do**

7:         **for** $i = 1:n$ **do**

8:             Initialize $b_{it} : b_{it} \leftarrow 0$

9:         **for** $r$ iterations **do**

10:            **for** $i = 1:n$ **do**

11:                 Generate $c_{it} : c_{it} \leftarrow \text{softmax}(b_{it})$

12:                 Generate $\mathbf{x}_t : \mathbf{x}_t \leftarrow \sum_i c_{it} \mathbf{u}_{it}$

13:            **for** $i = 1:n$ **do**

14:                 Update $b_{it} : b_{it} \leftarrow b_{it} + \mathbf{u}_{it} \cdot \mathbf{x}_t$

15:     **return** $\mathbf{x}_t$

16: **end procedure**

---

Overall, the 1D input sequence with length $d$, through the 1D CapsNet, is converted to $d$ high-level capsules. Each high-level capsule corresponds to a data point in the input sequence, representing the high-level features extracted from this data point. Subsequently, all high-level capsules are fed to a recurrent layer with $d$ cells, typically an LSTM layer, to generate predictions of the target variable.

## 3.3. LSTM

LSTM networks hold a significant position in time series forecasting and are often used as key components in forecasting models. An LSTM network consists of a sequence of LSTM cells, which, through a recurrent connection, enables information to flow from the previous steps to the current step, thereby capturing the temporal dependencies between data points at different positions in the input. The LSTM cell incorporates a memory mechanism to alleviate the vanishing gradient problem in simple RNNs (Hochreiter & Schmidhuber, 1997). It can store information for long durations, selectively forget irrelevant information, and update content based on a new input.

As shown in Figure 2, the LSTM cell architecture consists of four key components: forget gate $\mathbf{f}_t$, update gate $\mathbf{u}_t$, candidate state $\tilde{\mathbf{c}}_t$, and output gate $\mathbf{o}_t$. It receives three inputs at each time step: the current cell input $\mathbf{x}_t$ corresponding to the current time step $t$, the previous cell output $\mathbf{h}_{t-1}$, and the previous cell state $\mathbf{c}_{t-1}$. These inputs are processed using the internal gates of the LSTM cell, resulting in the corresponding output $\mathbf{h}_t$ and cell state $\mathbf{c}_t$. The outputs of these gates were calculated using the following equations:

$$\mathbf{f}_t = \sigma(\mathbf{W}_f \cdot [\mathbf{h}_{t-1}, \mathbf{x}_t] + \mathbf{b}_f) \tag{5}$$

$$\mathbf{u}_t = \sigma(\mathbf{W}_u \cdot [\mathbf{h}_{t-1}, \mathbf{x}_t] + \mathbf{b}_u) \tag{6}$$

$$\tilde{\mathbf{c}}_t = \tanh(\mathbf{W}_c \cdot [\mathbf{h}_{t-1}, \mathbf{x}_t] + \mathbf{b}_c) \tag{7}$$

$$\mathbf{c}_t = \mathbf{f}_t * \mathbf{c}_{t-1} + \mathbf{u}_t * \tilde{\mathbf{c}}_t \tag{8}$$

$$\mathbf{o}_t = \tanh(\mathbf{W}_o \cdot [\mathbf{h}_{t-1}, \mathbf{x}_t] + \mathbf{b}_o) \tag{9}$$

$$\mathbf{h}_t = \mathbf{o}_t * \tanh(\mathbf{c}_t), \tag{10}$$

where tanh is the hyperbolic tangent function and $\sigma$ is the sigmoid function. The symbol "$\cdot$" means the dot product of the matrices, and "$*$" is the elementwise multiplication. The matrices $\mathbf{W}_f$ $\mathbf{W}_u$ $\mathbf{W}_c$ $\mathbf{W}_o$ and $\mathbf{b}_f$ $\mathbf{b}_u$ $\mathbf{b}_c$ $\mathbf{b}_o$ are the gate weights and bias, respectively.

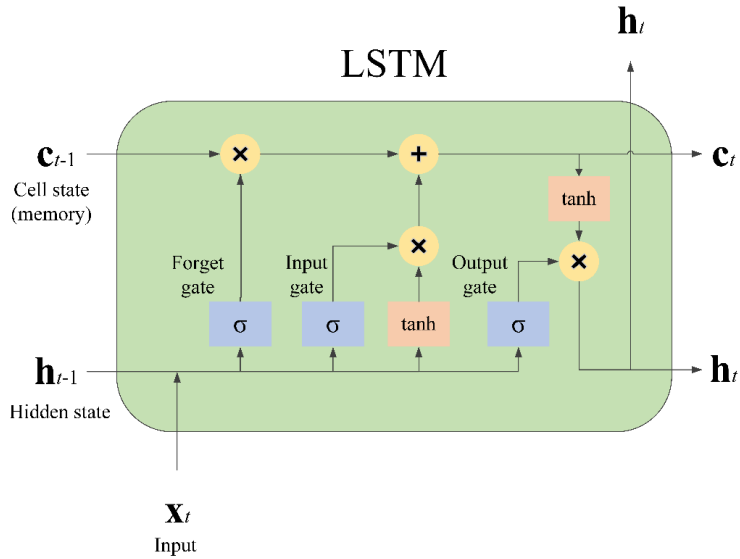

**Figure 2. The architecture of the LSTM cell.**

For multi-step time series forecasting, a "many-to-one" LSTM layer is employed, which uses a sequence of vectors as input and generates one vector output at the last time step. Figure 3 shows the LSTM layer used to process a sequence of vectors with $d$ time steps. Each vector of the input sequence is a high-level capsule generated from 1D CapsNet. The output of this LSTM layer is usually passed through a dense layer with a suitable activation function to obtain multi-step predicted values.

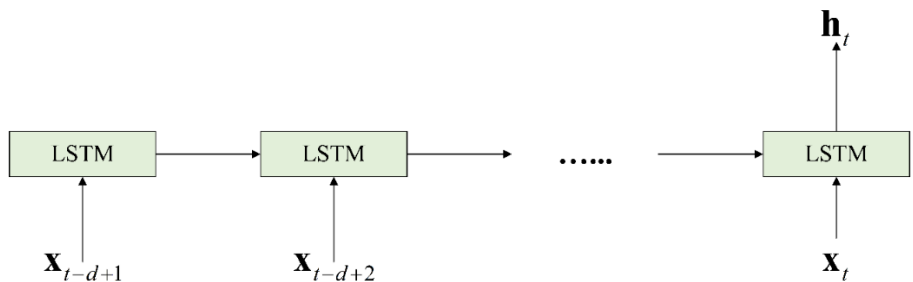

**Figure 3. The "many to one" LSTM layer.**

### 3.4. 1D-CapsNet-LSTM

The proposed 1D-CapsNet-LSTM model, as depicted in Figure 4, is a hybrid neural network that leverages the strengths of the multiple components mentioned above to effectively perform a multi-step forecasting task. It comprises two key components: a 1D CapsNet and an LSTM network, all working in concert to process 1D sequential data and predict multi-step future values of the target variable under the MIMO strategy.

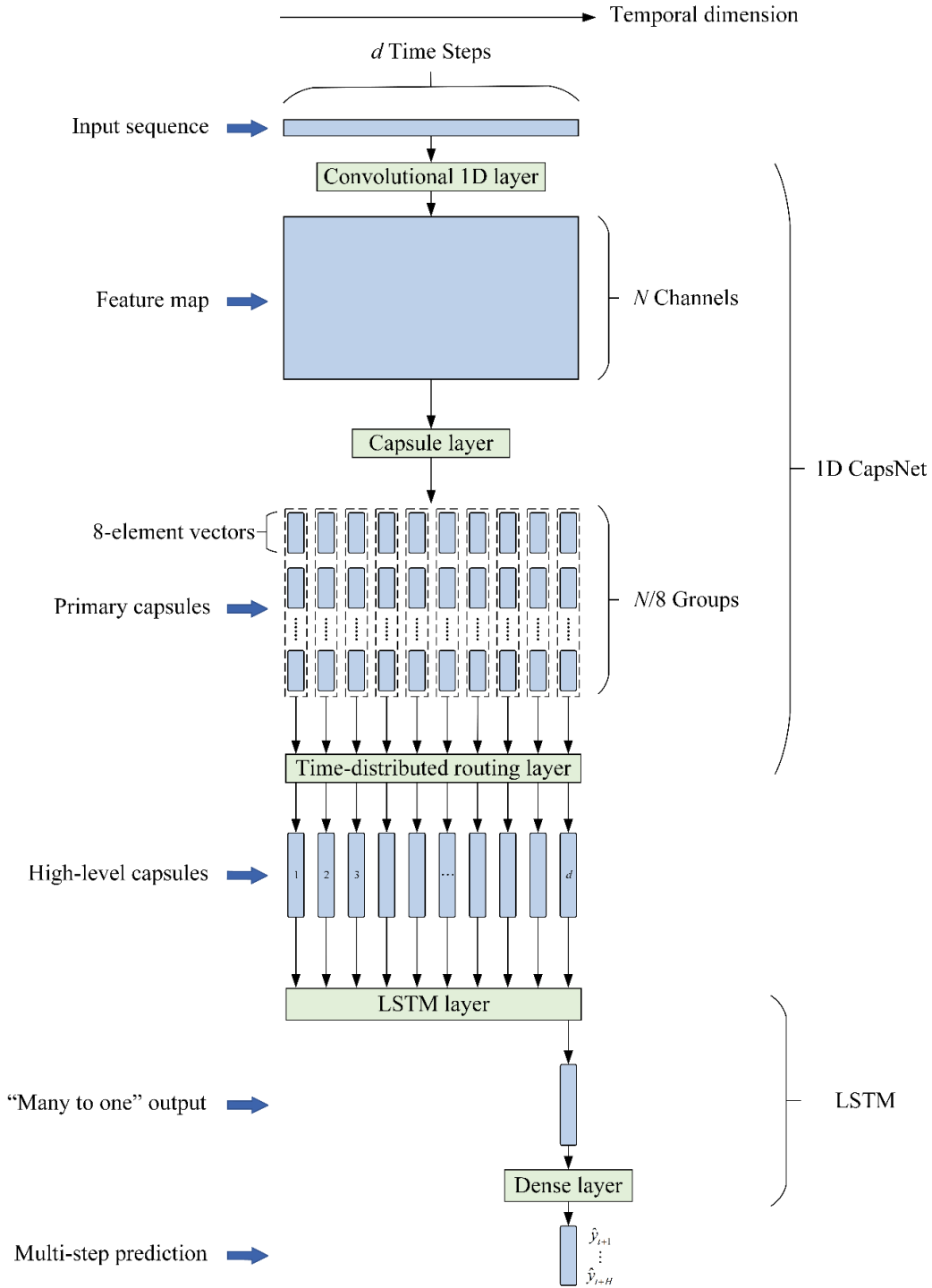

**Figure 4. The 1D-CapsNet-LSTM model for multi-step time series forecasting.**

Unlike traditional neural networks, which employ scalar neurons to represent learned features, CapsNet utilizes "capsules" to represent features in vector form, allowing for richer and more informative feature representations. This feature extraction approach is particularly beneficial when dealing with complex patterns and structures in sequential data. The LSTM network includes a "many to one" LSTM layer, which is responsible for capturing the temporal dependencies inherent in the high-level capsules generated by 1D CapsNet, and a dense (fully connected) layer, which maps the information extracted by the LSTM layer into a vector that represents the predicted values of the target variable at several consecutive future time steps. Through the 1D CapsNet and LSTM networks, meaningful predictions are obtained based on the learned features and temporal dependencies.

## 4. Experimental setup and performance comparison

Assessing the performance of the proposed 1D-CapsNet-LSTM model for multi-step forecasting of a stock index involves several steps. First, the raw data were extracted, cleaned, split, and normalized. A sliding window was then used to extract fixed-length segments of the data to construct the input sequences and labels. Second, the proposed and baseline models (LSTM, RNN, and CNN-SLTM) were constructed and then trained with input sequences and labels. Model performance was determined by comparing the predicted and actual values of the test set using a set of performance metrics. Third, the model performances were compared to demonstrate the effectiveness and superiority of the proposed model.

### 4.1. Data description and preprocessing

Four stock indices, including the S&P 500, DJIA, IXIC, and NYSE, were selected for the experiments. The raw data of these indices, covering ten years of historical daily close prices from January 1, 2010, to December 31, 2019, were retrieved from the Yahoo Finance website. The four univariate financial time series are visualized in Figure 5 and briefly described in Table 1.

For each stock index, the raw data were split into a training, validation, and test set, following an 8:1:1 ratio. This data splitting serves the purposes of training, fine-tuning, and evaluating models, ensuring their ability to generalize to new data while preventing overfitting to the training set. The training set consisted of 2014 observations, whereas the validation and test sets consisted of 251 observations. After data splitting, min–max normalization was performed on the training set to improve the convergence of the deep learning algorithms. The min–max normalization is given by Equation (11):

$$y_t^{normalized} = \frac{y_t - \min(y_t)}{\max(y_t) - \min(y_t)}, \tag{11}$$

where $y_t$ represents the data points in the training set, $y_t^{normalized}$ represents the normalized data points in the training set, $\min(y_t)$ is the minimum value of $y_t$, and $\max(y_t)$ is the maximum value of $y_t$. In addition, the values of $\min(y_t)$ and $\max(y_t)$ were used to normalize the validation and test sets.

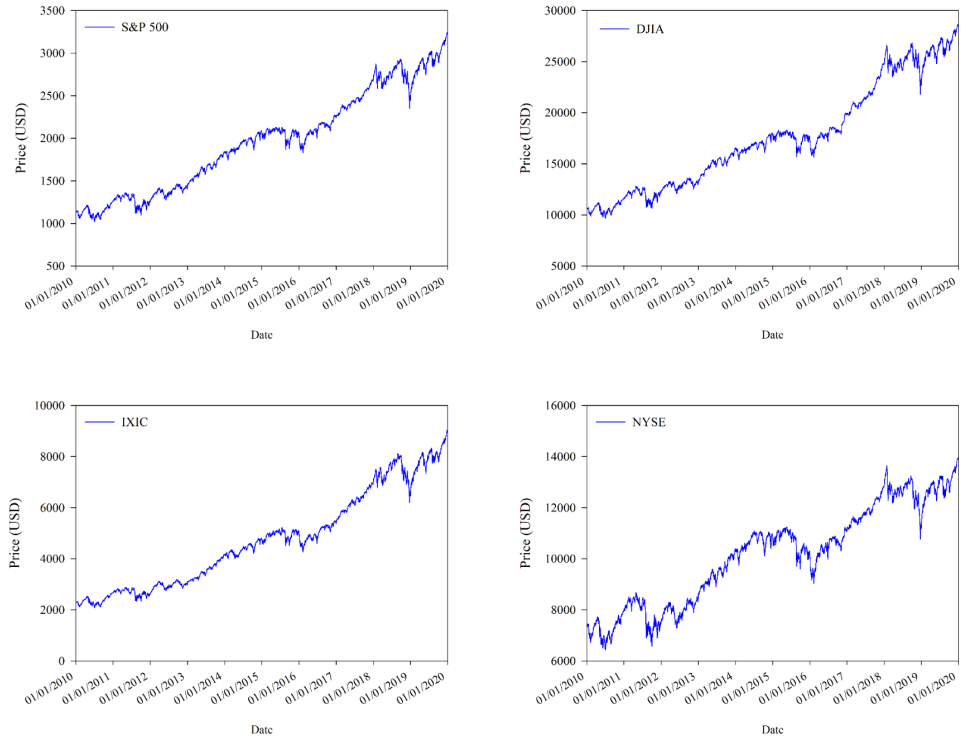

Figure 5. Close price series of four stock indices.

Table 1. Statistics of four close price series.

| Financial time series | Count | Mean | Standard Deviation | Minimum | Median | Maximum |
| --- | --- | --- | --- | --- | --- | --- |
| S&P 500 | 2516 | 1962.60886 | 588.91025 | 1022.58 | 1986.48 | 3240.02 |
| DJIA | 2516 | 17606.74157 | 5147.05011 | 9686.48 | 17008.23 | 28645.26 |
| IXIC | 2516 | 4744.1578 | 1878.80328 | 2091.79004 | 4620.54517 | 9022.38965 |
| NYSE | 2516 | 10162.17652 | 1942.52197 | 6434.81006 | 10440.30518 | 13944.13965 |

After data normalization, the entire time series was transformed into short sequences using a sliding window approach in a supervised learning scheme. Fixed-length segments of data were extracted as the window (time lag) was moved over the entire series. The segments from the training set and corresponding observations of the target variable construct pairs of sequences and labels for model training. The sliding window approach for preparing input sequences and labels is shown in Figure 6. In particular, the data segment length, or the input sequence length $d$, was set to 50 to efficiently process a context with sufficient information while avoiding excessive computational and memory requirements. The following five data points were regarded as the corresponding label of the input sequence, covering the forecasting horizon that would not lead to extremely high prediction error. This operation was conducted sequentially by shifting one time step to the future each time to produce all input sequences and labels. Labels were not needed for the validation and test sets, and only input sequences were prepared. The predictions provided by the forecasting model were denormalized using the same parameters given by Equation (11) to return the output to the original scale.

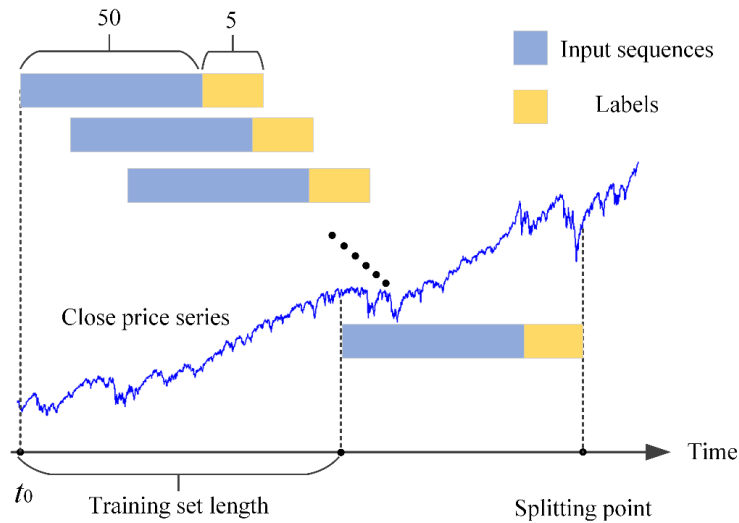

Figure 6. Sliding windows for preparing input sequences and labels.

### 4.2. Model configuration

To ensure that the difference in the models' performance is primarily due to the specific part of each neural architecture rather than other sources such as hyperparameter optimization, the common settings were used for the convolutional and LSTM layers in all models. The hyperparameters that needed tuning mainly refer to the dimension of high-level capsules, the iteration times of dynamic routing for each temporal slice, and the learning rate, as presented in Table 2. This study adopted the hyperband method for hyperparameter tuning, which can efficiently balance the exploration of different hyperparameter configurations with the allocation of computational resources, allowing for the discovery of optimal hyperparameters more quickly and cost-effectively than traditional grid search or random search methods (Li et al., 2017). During the hyperparameter tuning process, numerous configurations with different hyperparameters were randomly sampled and trained for a small number of epochs. After this initial phase, only the best-performing configurations were selected to proceed to the next round, where they were allocated more resources or training epochs. This process was continued until only one configuration remained. In addition, the number of training epochs was set to 400 to ensure that each model was properly trained and converged. The learning rate decreased by 5% if no improvement was observed in model performance after five epochs. The Adam optimizer (Kingma & Ba, 2014) was used to update the model parameters with the mean squared error (MSE) as the cost function. Table 3 provides an overview of the structures of the proposed and baseline models.

**Table 2. Hyperparameter tuning list.**

| Hyperparameter | Range |
|---|---|
| Dimension of high-level capsule | (256,512,768,1024) |
| Iteration times of dynamic routing for each temporal slice | (2,3,4,5) |
| Learning rate | (0.0001~0.01) |

**Table 3. Structure of forecasting models.**

| Forecasting model | Layer | Parameters | Output Shape | Parameter scale |
|---|---|---|---|---|
| CapsNet-LSTM | InputLayer | | (50, 1) | 0 |
| | Conv1D | filters = 256<br>kernel_size = 2<br>strides = 1 | (50, 256) | 768 |
| | Reshape | dimension of primary capsule = 8 | (50, 32, 8) | 0 |
| | Lambda (Squashing) | | (50, 32, 8) | 0 |
| | Time-distributed Routing | dimension of high-level capsule = 256<br>iteration times = 3 | (50, 256) | 65536 |
| | LSTM | hidden unit = 200 | (200) | 365600 |
| | Dense | | (5) | 1005 |
| LSTM | InputLayer | | (50, 1) | 0 |
| | LSTM | hidden unit = 200 | (200) | 161600 |
| | Dense | | (5) | 1005 |
| RNN | InputLayer | | (50, 1) | 0 |
| | SimpleRNN | hidden unit = 200 | (200) | 40400 |
| | Dense | | (5) | 1005 |
| CNN-LSTM | InputLayer | | (50, 1) | 0 |
| | Conv1D | filters = 256<br>kernel_size = 2<br>strides = 1 | (50, 256) | 768 |
| | MaxPooling1D | pool_size = 2<br>strides = 1 | (50, 256) | 0 |
| | LSTM | hidden unit = 200 | (200) | 365600 |
| | Dense | | (5) | 1005 |

The proposed and baseline models were implemented using Python TensorFlow and the high-level API Keras. In addition, a distributed strategy[1] using eight tensor processing units[2] (TPUs) was employed for model training. Under the distributed strategy, each batch of training data comprising 32 samples was divided into eight groups and distributed across eight TPUs for code execution. The gradients produced by each TPU were aggregated to update the model parameters.

### 4.3. Evaluation metrics

After model training, the performance of the proposed and baseline models was evaluated using four evaluation metrics, including the root mean squared error (RMSE), mean absolute error (MAE), mean absolute percentage error (MAPE), and Theil inequality coefficient (TIC), based on the prediction of the test set. The formulas for these metrics are as follows:

$$RMSE = \sqrt{\frac{1}{n}\sum_{i=1}^{n}(y_i - \hat{y}_i)^2} \tag{12}$$

$$MAE = \frac{1}{n}\sum_{i=1}^{n}|y_i - \hat{y}_i| \tag{13}$$

$$MAPE(\%) = \frac{100}{n}\sum_{i=1}^{n}\left|\frac{y_i - \hat{y}_i}{y_i}\right| \tag{14}$$

$$TIC = \frac{\sqrt{\frac{1}{n}\sum_{i=1}^{n}(y_i - \hat{y}_i)^2}}{\sqrt{\frac{1}{n}\sum_{i=1}^{n}y_i^2} + \sqrt{\frac{1}{n}\sum_{i=1}^{n}\hat{y}_i^2}}, \tag{15}$$

where $n$ denotes the test set size, $\hat{y}_i$ denotes the predicted value, and $y_i$ denotes the actual stock index value. The RMSE places greater emphasis on the highest error and is therefore more sensitive to outliers, whereas the MAE is more robust to outliers. The MAE and MAPE were used to determine the average difference between the predicted and actual values, and the TIC provided insight into how closely the estimated values tracked the actual values over time. In general, smaller values of these metrics indicate more accurate and reliable forecasts (Gilliland, 2010).

### 4.4. Performance comparison

This section compares the performance of all models based on the prediction of the test set in terms of the RMSE, MAE, MAPE, and TIC. Figures 7, 8, and 9 present the one-step-ahead, three-step-ahead, and five-step-ahead forecasting results for the four stock indices, respectively. In general, as the forecasting horizon extends, forecasting errors accumulate, resulting in a gradual decline in the forecasting accuracy. Through observations, the 1D-CapsNet-LSTM model exhibited improvements in prediction accuracy compared to the baseline models across all forecast horizons. This enhancement can be attributed to the remarkable feature extraction capabilities of the 1D CapsNet, as all models adopted the same settings for the common parts. Therefore, the 1D CapsNet plays a pivotal role in enhancing the performance of the proposed model in multi-step stock index forecasting. It is noteworthy that the RNN model delivers subpar forecasting results for the DJIA and NYSE stock indices. This suboptimal performance can be attributed to its simplistic structure and relatively fewer parameters than those of other models. This simplicity renders the RNN model more susceptible to overfitting, which leads to diminished model performance.

---
[1] https://www.tensorflow.org/tutorials/distribute/keras
[2] https://cloud.google.com/tpu/docs/system-architecture-tpu-vm

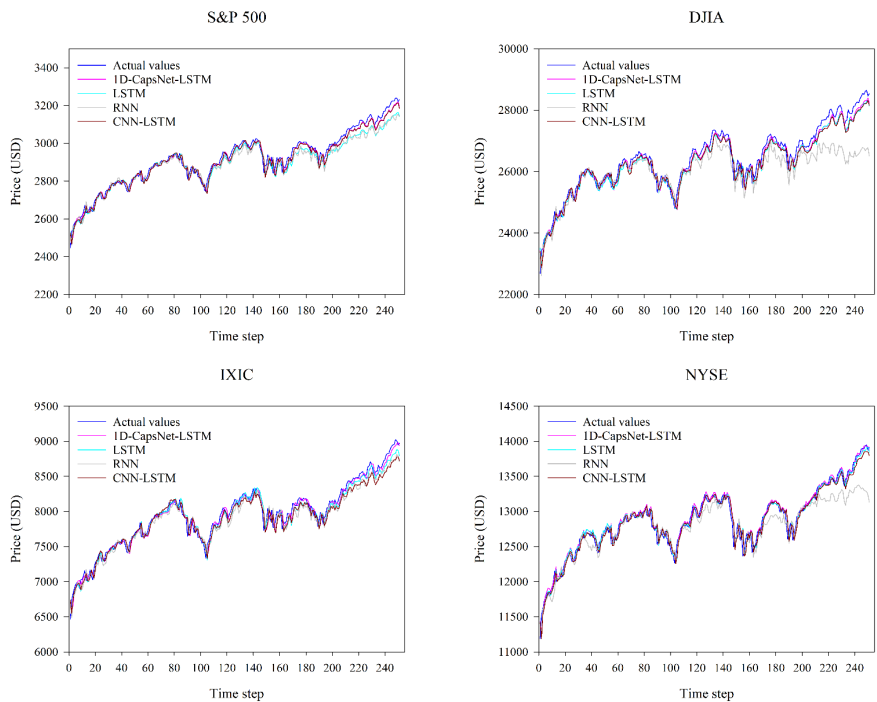

**Figure 7. One-step-ahead forecasting results of four stock indices.**

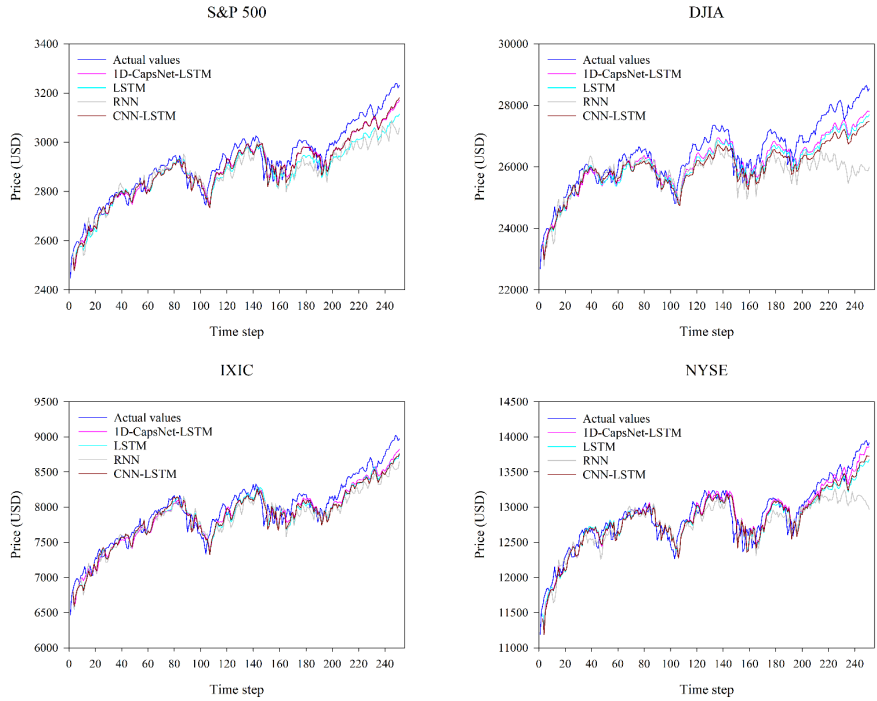

**Figure 8. Three-step-ahead forecasting results of four stock indices.**

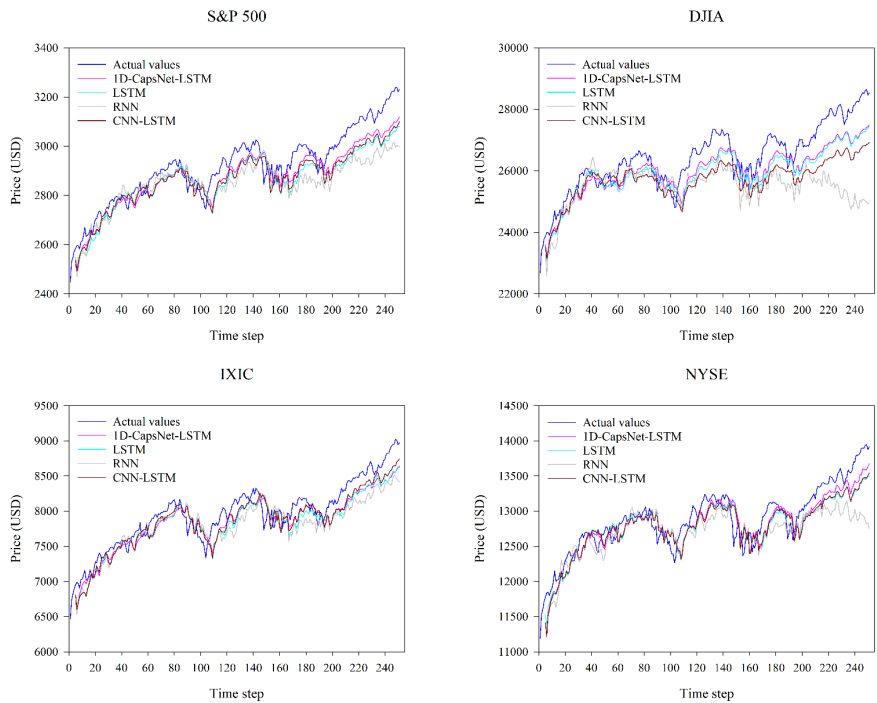

**Figure 9. Five-step-ahead forecasting results of four stock indices.**

Tables 4, 5, 6, and 7 provide the performance comparisons of all forecasting models in the five-step forecasting of the S&P 500, DJIA, IXIC, and NYSE, respectively. From Table 4, which presents all models' performance on S&P 500 forecasting, it is found that the 1D-CapsNet-LSTM model, with the lowest values of RMSE, MAE, MAPE, and TIC on one-step-ahead, three-step-ahead, four-step-ahead, and five-step-ahead forecasting, consistently outperformed other baseline models, demonstrating its superior predictive accuracy. In contrast, the RNN and LSTM models tend to exhibit higher prediction errors, particularly when the forecasting horizon is long. The CNN-LSTM model performed competitively, and its performance fell between that of the 1D-CapsNet-LSTM and LSTM models. It was also observed that for the proposed model, the forecast error was higher over a long horizon than over a short horizon, and the four-step-ahead forecast error was approximately twice that of the one-step-ahead forecast. Similar trends are observed in Tables 5, 6, and 7. The proposed 1D-CapsNet-LSTM consistently outperformed the other models in terms of all evaluation metrics, except for a few cases in which LSTM and CNN-LSTM performed better for one or two specific forecast horizons. This finding implies that the 1D-CapsNet-LSTM demonstrated superior performance in most cases. Nonetheless, it is worth highlighting that stock market indices are highly volatile and stochastic and that a single type of deep learning model is unlikely to generate accurate predictions across all scenarios.

Table 4. Performance comparison of different models for S&P 500 forecasting.

| Forecasting horizon | Models | RMSE | MAE | MAPE | TIC |
|---|---|---|---|---|---|
| 1-step ahead | 1D-CapsNet-LSTM | **25.2** | **19.61** | **0.68** | **0.00432** |
| | RNN | 40.39 | 32.09 | 1.08 | 0.00695 |
| | LSTM | 38.13 | 30.42 | 1.03 | 0.00656 |
| | CNN-LSTM | 26.52 | 20.67 | 0.71 | 0.00455 |
| 2-step ahead | 1D-CapsNet-LSTM | 36.6 | 30.68 | 1.04 | 0.00629 |
| | RNN | 59.44 | 47.82 | 1.61 | 0.01024 |
| | LSTM | 43.14 | 36.41 | 1.23 | 0.00742 |
| | CNN-LSTM | **35.17** | **29.1** | **0.99** | **0.00604** |
| 3-step ahead | 1D-CapsNet-LSTM | **45.74** | **38.77** | **1.32** | **0.00786** |
| | RNN | 74.18 | 59.49 | 2 | 0.0128 |
| | LSTM | 64.15 | 52.91 | 1.78 | 0.01106 |
| | CNN-LSTM | 46.31 | 38.83 | 1.32 | 0.00796 |
| 4-step ahead | 1D-CapsNet-LSTM | **55.05** | **46.61** | **1.58** | **0.00947** |
| | RNN | 96.69 | 76.75 | 2.56 | 0.01673 |
| | LSTM | 63.18 | 54.03 | 1.83 | 0.01089 |
| | CNN-LSTM | 62.41 | 52.69 | 1.78 | 0.01076 |
| 5-step ahead | 1D-CapsNet-LSTM | **64.11** | **54.31** | **1.83** | **0.01104** |
| | RNN | 103.25 | 84.36 | 2.82 | 0.01787 |
| | LSTM | 81.85 | 69.29 | 2.33 | 0.01414 |
| | CNN-LSTM | 74.95 | 63.27 | 2.13 | 0.01294 |

**Table 5. Performance comparison of different models for DJIA forecasting.**

| Forecasting horizon | Models | RMSE | MAE | MAPE | TIC |
|---|---|---|---|---|---|
| 1-step ahead | 1D-CapsNet-LSTM | **233.07** | **186.4** | **0.71** | **0.00442** |
| | RNN | 636.5 | 446.09 | 1.65 | 0.01214 |
| | LSTM | 252.5 | 207.25 | 0.78 | 0.00479 |
| | CNN-LSTM | 254.62 | 208.06 | 0.79 | 0.00483 |
| 2-step ahead | 1D-CapsNet-LSTM | **306.97** | **256.52** | **0.97** | **0.00582** |
| | RNN | 796.88 | 583.24 | 2.16 | 0.01522 |
| | LSTM | 338.98 | 292.92 | 1.1 | 0.00644 |
| | CNN-LSTM | 443.43 | 374.45 | 1.4 | 0.00844 |
| 3-step ahead | 1D-CapsNet-LSTM | **461.87** | **392.13** | **1.47** | **0.00879** |
| | RNN | 963.34 | 715.47 | 2.64 | 0.01844 |
| | LSTM | 522.32 | 444.12 | 1.66 | 0.00995 |
| | CNN-LSTM | 611.06 | 512.39 | 1.91 | 0.01166 |
| 4-step ahead | 1D-CapsNet-LSTM | **582.55** | **493.69** | **1.84** | **0.0111** |
| | RNN | 1183.11 | 871.35 | 3.21 | 0.02271 |
| | LSTM | 606.5 | 520.32 | 1.94 | 0.01156 |
| | CNN-LSTM | 759.05 | 639.93 | 2.38 | 0.01451 |
| 5-step ahead | 1D-CapsNet-LSTM | **617.14** | **524.68** | **1.96** | **0.01176** |
| | RNN | 1343.31 | 1019.67 | 3.77 | 0.02585 |
| | LSTM | 678.45 | 581.09 | 2.17 | 0.01295 |
| | CNN-LSTM | 898.95 | 760.34 | 2.82 | 0.01722 |

**Table 6. Performance comparison of different models for IXIC forecasting.**

| Forecasting horizon | Models | RMSE | MAE | MAPE | TIC |
|---|---|---|---|---|---|
| 1-step ahead | 1D-CapsNet-LSTM | **80.89** | **61.72** | **0.79** | **0.00509** |
| | RNN | 113.67 | 92.54 | 1.16 | 0.00717 |
| | LSTM | 90.23 | 71.58 | 0.9 | 0.00568 |
| | CNN-LSTM | 107.75 | 86.02 | 1.07 | 0.00679 |
| 2-step ahead | 1D-CapsNet-LSTM | **112.36** | **92.01** | **1.16** | **0.00707** |
| | RNN | 168.47 | 140.34 | 1.74 | 0.01064 |
| | LSTM | 117.42 | 96.25 | 1.21 | 0.00739 |
| | CNN-LSTM | 144.39 | 117.64 | 1.46 | 0.00911 |
| 3-step ahead | 1D-CapsNet-LSTM | **140.99** | **120.01** | **1.5** | **0.00889** |
| | RNN | 195.03 | 163.73 | 2.03 | 0.01233 |
| | LSTM | 160.01 | 134.17 | 1.67 | 0.0101 |
| | CNN-LSTM | 155.74 | 129.76 | 1.62 | 0.00983 |
| 4-step ahead | 1D-CapsNet-LSTM | **168.32** | **141.57** | **1.76** | **0.01062** |
| | RNN | 228.19 | 191.61 | 2.37 | 0.01444 |
| | LSTM | 185.4 | 155.69 | 1.94 | 0.01171 |
| | CNN-LSTM | 182.07 | 152.5 | 1.89 | 0.0115 |
| 5-step ahead | 1D-CapsNet-LSTM | 194.44 | 163.1 | 2.02 | 0.01228 |
| | RNN | 261.52 | 221.36 | 2.74 | 0.01656 |
| | LSTM | 213.25 | 180.02 | 2.24 | 0.01348 |
| | CNN-LSTM | **183.85** | **155.51** | **1.94** | **0.0116** |

**Table 7. Performance comparison of different models for NYSE forecasting.**

| Forecasting horizon | Models | RMSE | MAE | MAPE | TIC |
|---|---|---|---|---|---|
| 1-step ahead | 1D-CapsNet-LSTM | 91.37 | 66.41 | 0.52 | 0.00354 |
|  | RNN | 188.79 | 135.2 | 1.03 | 0.00736 |
|  | LSTM | **90.09** | **66.03** | **0.52** | **0.0035** |
|  | CNN-LSTM | 93.34 | 70 | 0.55 | 0.00363 |
| 2-step ahead | 1D-CapsNet-LSTM | **122.63** | **94.61** | **0.74** | **0.00476** |
|  | RNN | 230.59 | 175.37 | 1.34 | 0.00899 |
|  | LSTM | 131.95 | 106.33 | 0.83 | 0.00513 |
|  | CNN-LSTM | 132.75 | 106.56 | 0.83 | 0.00516 |
| 3-step ahead | 1D-CapsNet-LSTM | **148.2** | **117.02** | **0.91** | **0.00576** |
|  | RNN | 260.7 | 197.17 | 1.51 | 0.01017 |
|  | LSTM | 167.18 | 136.84 | 1.06 | 0.0065 |
|  | CNN-LSTM | 162.07 | 131.52 | 1.02 | 0.0063 |
| 4-step ahead | 1D-CapsNet-LSTM | **179.61** | **146.25** | **1.14** | **0.00698** |
|  | RNN | 310.07 | 232.95 | 1.78 | 0.0121 |
|  | LSTM | 203.61 | 165.93 | 1.28 | 0.00792 |
|  | CNN-LSTM | 197.17 | 160.46 | 1.24 | 0.00767 |
| 5-step ahead | 1D-CapsNet-LSTM | **203.55** | **169.35** | **1.31** | **0.00792** |
|  | RNN | 337.03 | 257.51 | 1.97 | 0.01316 |
|  | LSTM | 228.88 | 190.1 | 1.47 | 0.00891 |
|  | CNN-LSTM | 225.41 | 188.11 | 1.45 | 0.00878 |

Figure 10 presents a visual comparison of the RMSE values for the multi-step forecasting of the four stock indices using different models. Although all the RMSE values increased as the forecast horizon increased, implying the deterioration of forecasting accuracy, the proposed 1D CapsNet-LSTM model still outperformed the baseline models in two aspects. First, substantial decreases in the RMSE values were observed when the proposed and the baseline models were compared. For instance, in the five-step S&P 500 forecasting, comparing the 1D-CapsNet-LSTM with the LSTM model, the RMSE values decreased by 33.9%, 15.1%, 28.8%, 12.9%, and 21.7% for each forecasting horizon. In contrast, comparing the 1D-CapsNet-LSTM with the CNN-LSTM model, the RMSE values decreased by 5.0%, -4.1%, 1.2%, 11.8%, and 14.5%, respectively. Similarly, in the five-step DJIA forecasting, comparing the 1D-CapsNet-LSTM with the LSTM model, the RMSE values decreased by 7.7%, 9.4%, 11.6%, 3.9%, and 9.0%. In contrast, comparing the 1D-CapsNet-LSTM with the CNN-LSTM model, the RMSE values decreased by 8.5%, 30.8%, 24.4%, 23.2%, and 31.3%, respectively. Similar phenomena were observed for the remaining forecasting tasks. Second, as the forecast horizon increased, the RMSE values of the 1D-CapsNet-LSTM model increased at a slower rate than those of the baseline models. When the forecasting horizon was short, the disparity in the RMSE values between the 1D-CapsNet-LSTM and baseline models was relatively modest in comparison with the disparity observed when dealing with longer forecasting horizons. This result indicates that the 1D-CapsNet-LSTM is more robust for multi-step forecasting. A comparison of the other metric values also showed a similar trend.

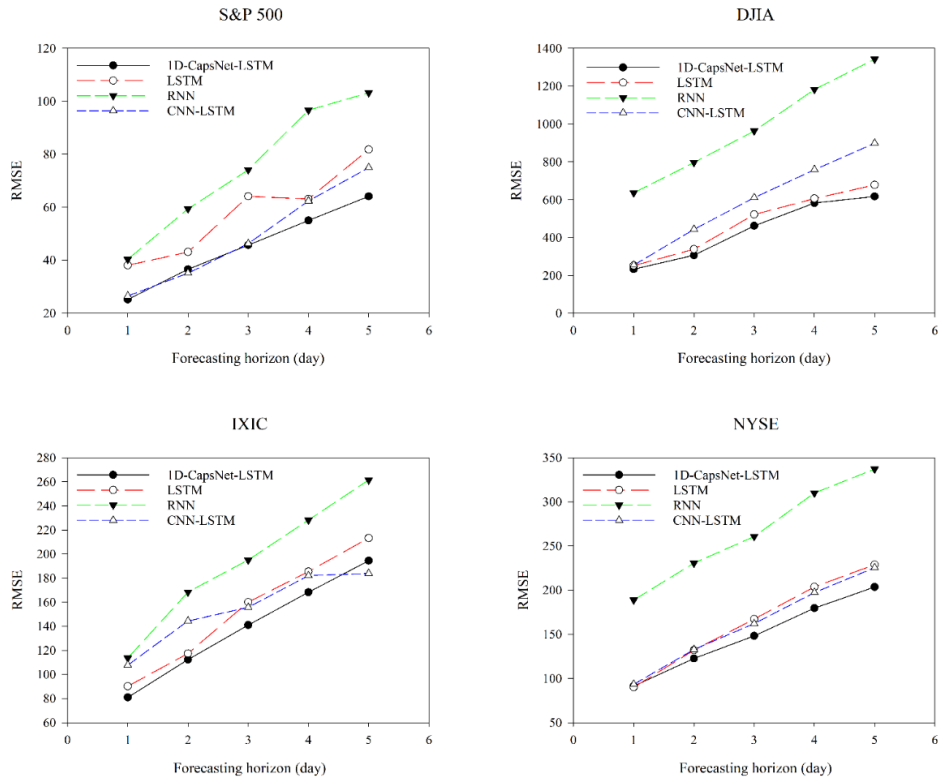

Figure 10. Comparison of RMSE values for five-step forecasting.

Based on the experimental results, we made the following key observations. First, the forecast errors tend to increase as the forecasting horizon increases, regardless of the forecasting models. In other words, it is generally more challenging to make accurate predictions for longer time horizons. This is a common phenomenon in financial time series forecasting, suggesting that uncertainty in predictions increases over time. In particular, the four-step-ahead forecast error is approximately twice as large as the one-step-ahead forecast error. This indicates that the error does not increase linearly with the forecasting horizon; instead, it exhibits a certain level of exponential or nonlinear growth. Second, multi-step forecasts are often subject to more complex and unpredictable factors, which can make accurate predictions more difficult. Factors such as changing market conditions, economic shifts, and unforeseen events may have a greater impact on multi-step predictions. This observation highlights the importance of selecting appropriate forecasting models and methods, particularly for long forecasting horizons. Models that perform well for single-step predictions may not necessarily perform well for multi-step predictions, and vice versa.

Overall, the proposed 1D-CapsNet-LSTM model outperformed the baseline deep learning models for multi-step stock index forecasting in terms of RMSE, MAE, MAPE, and TIC. The 1D-CapsNet-LSTM model achieved accurate forecasting results for one-step-ahead forecasting while maintaining stable performance in multi-step-ahead forecasting as the forecasting horizon increased; therefore, it is a reliable and robust option for similar forecasting tasks.

### 4.5. Practical considerations

Apart from evaluating the accuracy of the forecasts given by different models, some practical aspects should be considered when implementing the 1D-CapsNet-LSTM model for real-world applications. First, the training speed of the 1D-CapsNet-LSTM model is a crucial factor in determining its suitability for a specific task. Training speed was measured by recording and comparing the time taken by each model to train a batch of training sets. The problem of implementing a 1D-CapsNet-LSTM model often involves a longer training time than a CNN-LSTM model because the nested routing operation in the 1D CapsNet is slower than the pooling operation in the CNN (Ma et al., 2021). Nonetheless, the distributed training strategy is effective in reducing the training time of the 1D-CapsNet-LSTM model.

Figure 11 shows a comparison of the training speeds of the proposed and baseline models. For one batch of data, the training time of the 1D-CapsNet-LSTM model improved from approximately 1000 ms to 30 ms, which was much closer to the training time of the baseline models. The use of a distributed strategy enhanced the ability of the 1D-CapsNet-LSTM model to effectively handle large-scale prediction tasks. It is worth noting that in this study, the training of the 1D-CapsNet-LSTM model was accelerated with the aid of the parallel processing power offered by the TPUs. In addition, 1D-CapsNet-LSTM models can be deployed on field-programmable gate arrays (FPGAs), which can speed up model training through massive parallel computation while consuming less energy than graphics processing units (GPUs) or TPUs.

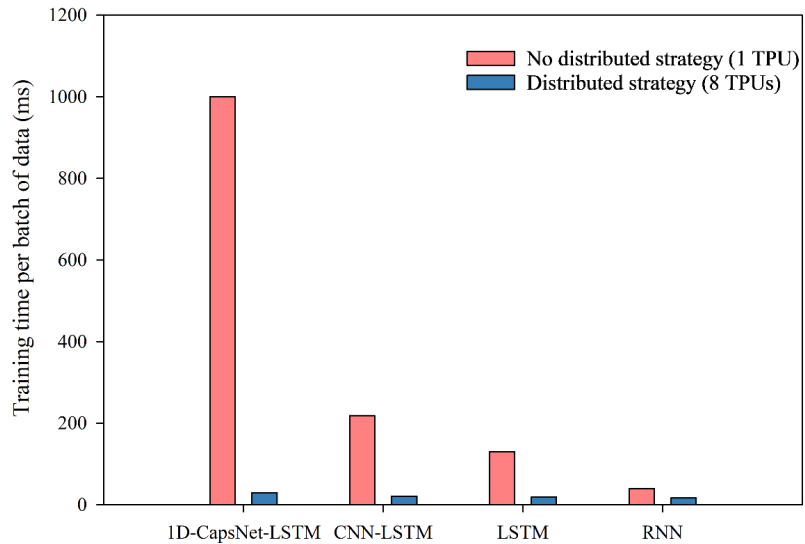

**Figure 11. Training speed comparison.**

Second, although deep learning models have the potential to enhance traders' understanding of market behavior by analyzing vast amounts of data, identifying complex patterns, and providing a competitive edge over traditional methods, integrating these models into real-time trading systems remains challenging because of factors such as model complexity, the need for sufficient training data, and the need to adapt to changing market conditions. Additional research and development are necessary to effectively incorporate the 1D-CapsNet-LSTM model into high-frequency trading systems.

## 5. Conclusion

Accurately predicting the multi-step future prices of a stock market index is crucial for profitable trading, risk management, and informed investment decision making. However, forecasting results are often unsatisfactory owing to the stochastic and volatile nature of the data. Researchers have made various attempts, and this process is ongoing. Inspired by CNN-LSTM networks, which employ a 1D CNN as a sophisticated feature extraction component to improve model performance, this study aims to investigate the potential of CapsNet as a more advanced feature extraction component in the LSTM-based forecasting model to improve the multi-step forecasting result. In this study, we propose a hybrid deep learning model, 1D-CapsNet-LSTM, that integrates the 1D CapsNet to extract high-level capsules from the 1D sequence and the LSTM layer to capture the temporal dependencies between these capsules.

Under the MIMO strategy, the performance of the proposed 1D-CapsNet-LSTM model was evaluated based on five-step forecasting of four real-world stock market indices, the S&P 500, DJIA, IXIC, and NYSE, using four evaluation metrics, RMSE, MAE, MAPE, and TIC. Through a performance comparison of the proposed model and baseline models, including LSTM, RNN, and CNN-LSTM, it was found that the proposed model achieved accurate results for one-step-ahead forecasting and exhibited the most stable performance in multi-step-ahead forecasting for different forecast horizons. This suggests that the 1D-CapsNet-LSTM model is a more reliable and robust option for multistep-ahead forecasting tasks than other deep learning models.

Although the 1D-CapsNet-LSTM network has a complex architecture, it still shows great promise for large-scale and complex prediction tasks because its training time can be significantly reduced through a distributed training strategy using eight TPUs. In addition, with the advancement of computing hardware and software technologies, the 1D-CapsNet-LSTM architecture can be deployed on FPGAs, which are energy-saving and enable efficient model training through massive parallel computations.

Furthermore, it is important to note that the model evaluation was based solely on stock index data; therefore, additional studies are needed to determine the effectiveness of the 1D-CapsNet-LSTM network in other domains. Finally, the proper hyperparameter setting of the 1D-CapsNet-LSTM is worth further investigation. The 1D CapsNet designed in this study is conceptualized based on the assumption that there is a "one-to-one correspondence" relationship between the original data points from the input sequence and high-level capsules. This assumption leaves space for future work to explore the optimal setting of this relationship. Whether there should be a "many-to-one" relationship and how we should select the corresponding temporal slices for generating one high-level capsule are challenging research questions that should be answered in the future.

**Acknowledgment**

This study did not receive any specific grants from funding agencies in the public, commercial, or not-for-profit sectors.

**Declaration of competing interest**

The authors declare that they have no competing financial interests or personal relationships that could influence the study reported herein.

**Author contributions**

**Cheng Zhang**: Conceptualization; Data curation; Methodology; Resources; Software; Visualization; Writing - original draft; Writing – review & editing. **Nilam Nur Amir Sjarif**: Validation; Writing – review & editing; Supervision. **Roslina Ibrahim**: Writing – review & editing; Supervision.